# Resource Production of Written Forms of Sign Languages by a User-Centered Editor, SWift (SignWriting improved fast transcriber)

**Fabrizio Borgia**[1-2], **Claudia S. Bianchini**[3-4-5], **Patrice Dalle**[1], **Maria De Marsico**[2]

(1) Université "Paul Sabatier" de Toulouse (118, route de Narbonne, Toulouse)
(2) Dip. Informatica - Università degli Studi "La Sapienza" di Roma (Via Salaria 113, Roma)
(3) Université Paris 8 ED-CLI / CNRS-UMR7023-SFL (2, rue de la Liberté, Saint-Denis)
(4) Università degli Studidi Perugia (Piazza Morlacchi 1, Perugia)
(5) CNR-ISTC-SLDS (Via Nomentana 56, Roma)
Contact person: fabrizio.borgia@uniroma1.it

**Abstract**

The SignWriting improved fast transcriber (SWift), presented in this paper, is an advanced editor for computer-aided writing and transcribing of any Sign Language (SL) using the SignWriting (SW). The application is an editor which allows composing and saving desired signs using the SW elementary components, called "glyphs". These make up a sort of alphabet, which does not depend on the national Sign Language and which codes the basic components of any sign. The user is guided through a fully automated procedure making the composition process fast and intuitive. SWift pursues the goal of helping to break down the "electronic" barriers that keep deaf people away from the web, and at the same time to support linguistic research about Sign Languages features. For this reason it has been designed with a special attention to deaf user needs, and to general usability issues. The editor has been developed in a modular way, so it can be integrated everywhere the use of the SW as an alternative to written "verbal" language may be advisable.

## 1. Introduction

More or less 0.1% of the worlds population (according to the World Health Organisation), is deaf or hard of hearing. Most of the deaf use, as their favorite tongue, a Sign Language (SL)

Until 1960, SLs were considered a mimic form of communication, which could never reach the status of language. Stokoe's works (1960) were a starting contribution toward recognizing the richness and expressiveness of SLs, and their "linguisticity", i.e. the fact of being true languages in themselves, with very complex features, often so different as not being found in VLs. However, these works were all focused on the manual components of SL, decomposing signs in hand configuration, orientation, movement and location.

By the end of the '90, Cuxac's works (1996, 2000) on the iconicity in LSF (French Sign Language) added a new dimension on SL studies, focusing the attention on the non-manual components of signs. Cuxac demonstrated that most information in SL is transmitted through eye gaze, facial expressions and body movements. Those researches were then adapted to others SLs, e.g. by the Antinoro Pizzuto's research on LIS (Italian SL) (Antinoro Pizzuto, 2008; Antinoro Pizzuto *et al*, 2007; Pizzuto & Rossini, 2007)

Even being true languages, SLs, similarly to many other world languages, still have not developed their own writing system. Although it is possible to use a phonetics mean of transcription - like the IPA (International Phonetic Alphabet) - for vocal languages (VL), the typological specificity of SW does not permit similar solutions (Garcia & Boutet, 2006). This has deep consequences because, despite advances in the comprehension of patterns and rules underlying SLs, the lack of a system to transcribe them still hinders their study: *"That which one cannot write down, one cannot research"* (Boyes Braem, 2012). The search for a suitable transcription instrument is strongly connected to the lack of a writing system for SLs: in fact, most transcription system for VL are based on a preexistent writing form (eg., IPA is based on the latin alphabet). Thus, it is clear that "working at a writing system for SL is now the best way to make progress in SL transcription" (Garcia, 2007); moreover, researches believe that having means to represent SLs may be of paramount importance to study their structure and patterns in a deeper way (Antinoro Pizzuto *et al.*, 2010b;Pizzuto *et al.*, 2006;Di Renzo *et al.*, 2010;Pizzuto & Pietrandrea, 2001, Garcia 2007, Garcia & Derycke, 2011). The so-called "glosses" are a very common way of representing SLs using the written form of a verbal language. Glosses are labels in VL that express (in a very simplified way) the meaning of SL: we consider this kind of representation as a "pseudo-notation" (Bianchini, 2012) because it is a "trick" to bypass the notation, and to annotate the meaning of signs yet loosing the information of their form [see (Antinoro Pizzuto *et al.*,2010b; Garcia & Boutet 2007) for a critic]. There are also true notation forms, such as the Stokoe Notation System (Stokoe, 1960), or HamNoSys (Prillwitz, 1989), but none of them takes into consideration the non-manual expressions that are crucial in any SL. Information is also conveyed through glance, hands, facial expression, head and shoulders in a multi-linear way, therefore it is very difficult to transcribe an expression in any SL using said notations. These limitations on the component representation do not allow to use this systems to write signs that are not isolated and in citation form. In other words, it is impossible to transcribe real SL discourses.

Besides hindering research, the lack of a writing or transcription system for SL also limits the possibility to provide information (e.g. on the web) directly in a form "equivalent" to the signed content. This represents a serious accessibility flaw, since deaf people often have problems in acquiring and using written VLs, and this is a further obstacle in providing information in easily comprehensible ways to them (Antinoro Pizzuto *et al.*, 2010a, Garcia & Perini 2010). As a consequence, a deaf person who wants to venture into the digital world must overcome numerous and difficult barriers, just like those he faces in everyday life. As an example, while surfing through Internet, each user exploits semantic traces to judge which item or which site to choose. The problem, in the case of deaf people, is that these semantic traces are generally available (typically the text of the links) in a language that is not their native language, and in which they often have insufficient reading competence; in fact, some studies have shown (Fajardo *et al.*, 2008) that deaf users find it difficult to devise a strategy to gather the information they need through textual traces. It is worth reminding at this point that, according to the World Health Organization, 278 million people worldwide are deaf or have hearing difficulties, and many of them use SL as their mother tongue. "*The diversity of regional variations of SL constitutes a set of minority languages relatively underrepresented in the digital world. Thus, members of the deaf community usually face non-native language web sites where accessibility barriers may emerge*" (Fajardo *et al.*, 2009).

## 2. SignWriting

SignWriting (SW; Sutton, 1995), which will be discussed further on and to which the SWift project is addressed, seems at present the best solution to the problem of SL representation (Bianchini *et al.*, 2010; Bianchini, 2012; Gianfreda *et al.*, 2009, Di Renzo *et al.*, 2009 ). Its strength is in the ability to combine a bi-dimensional picture with a coding system which relies on highly iconic symbols, thus intuitive and easy to remember (Bianchini, 2012). In more detail, SW is a graphical framework which uses visual symbols (called glyphs) to represent configurations, movements and facial expressions making up SLs. It is part of a more complex writing system: the Sutton Movement Writing & Shorthand, devised by Valerie Sutton, an American choreographer, with the goal of providing notations to trace any body movement.

Fig. 1 shows a sign written in SW and illustrates examples of glyphs for (from top to bottom) facial expression (green), shoulders position (red), hand configuration (black), contact (blue, between the hands) and movement (pink). Usually SW is written only in black and white.

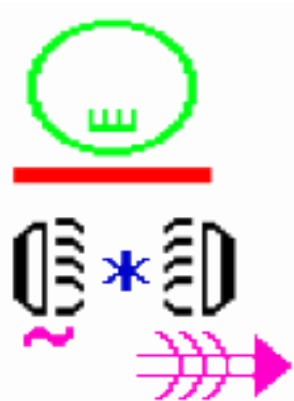

Fig. 1 – The LIS sign for "various" written using SW

One of the features that make SW very promising compared with other notations, is that it can express by itself a signed sequence, without further annotations in a written VL.

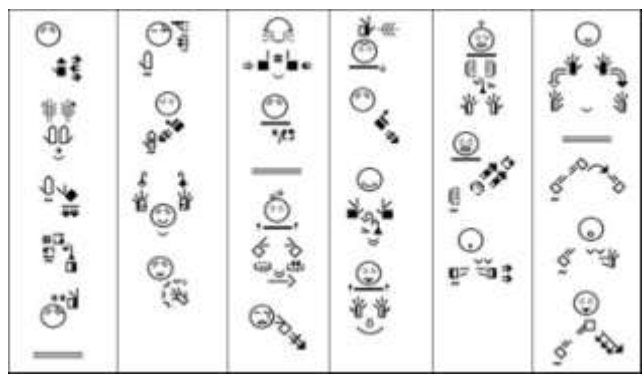

Fig. 2 - Example of SW text (source: Di Renzo *et al.*, 2012, segment of the "Indian hat" story). The text lays vertically, from top to bottom and left to right.

## 3. SignWriting Editors

The set of glyphs may be roughly considered as an alphabet that can be used to write any SL in the world; this set makes up the International SignWriting Alphabet (ISWA[1]). The latter is available as an archive of more than 35,000 images (.png), each representing a single glyph. Due to their huge number, glyphs have been coded by associating to each of them a string of decimals, with labels associated with each numeric component. Labels are used to group the glyphs according to a number of features (e.g. expressions of movements of the represented body part) to ease their handling.

Every two years, on average, ISWA evolves and integrates new glyphs: this causes a change in the coding of glyphs that create confusion between old and new versions.

A re-classification of ISWA was attempted by Bianchini (Bianchini, 2012), with the aim of improving its consistency and effectiveness. Within such re-classification, ISWA was divided in categories, families and sub-families, which contain prototypical glyphs that are "declined" according to rules without exceptions: each glyph is the unique combination of a prototype (from a category), a family and a sub-family, and an implementation of each applicable rule. This new arrangement of ISWA allows users a better understanding of the characteristics governing SW, making SW easier to learn than with the old classification.

---

[1] for more information see:
<http://www.movementwriting.org/symbolbank/>

In practice, SW is a system devised to allow deaf people as well as linguists to exploit an intuitive and easy-to-grasp written form of (any) sign language.

The research team of Valerie Sutton has developed SignMaker[2], an application that allow writing and saving signs using just a web browser. (Fig. 3).

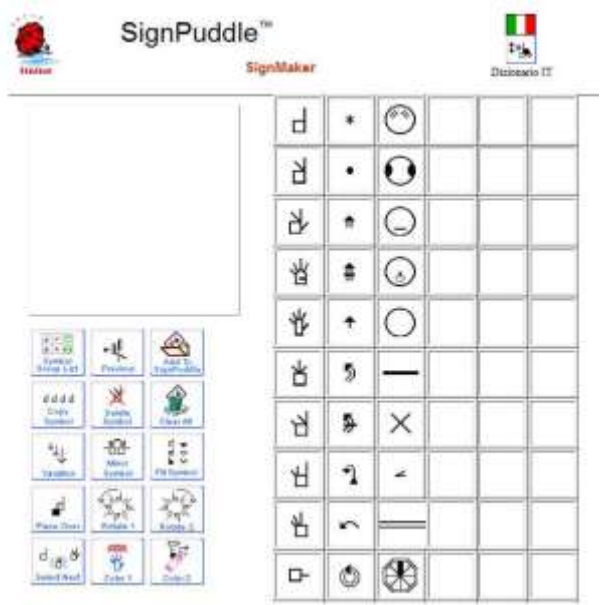

Fig. 3 – Home screen of SignMaker.

The expressiveness of the overall interface is quite good, some interface elements give an idea of the function to which they are associated, but all of them can be fully understood only by a small group of people: signing people with a good knowledge of SignWriting (Borgia, 2010). This way, the set of potential SignMaker users narrows.

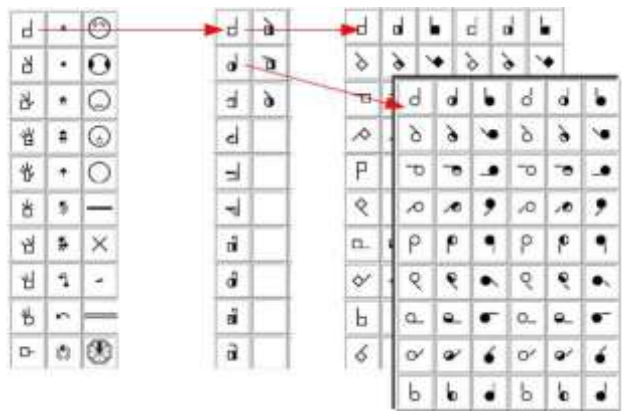

Fig. 4 – SignMaker's Glyph Menu: note its tree-like structure.

## 4. SWift

The interface of SWift appears more friendly than that of SignMaker: it minimizes the use of text labels and presents a collection of colorful and familiar icons (Fig. 5). The goal is to make the user feel comfortable and to avoid confusing her/him with a large amount of information. We realized that even a very careful redesign of the interface could not have, by itself, improved an application like the SignMaker. In fact, the redesign of the graphics should have supported a further complete redesign of the logic part of the application.

The interface of the brand-new application SWift was designed according to some prerequisites:
• Intuitive interface: the user does not need to "learn" to use the interface, he should rather "understand" it; for this purpose, each function is presented in intuitive and familiar way.
• Minimization of information: each interface screen only presents the necessary amount of information, to avoid confusing the user.
• Evocative icons: the icons are simple, familiar and large. If their meaning is not immediately understood, mouseover-activated animations can be started.
• Minimization of text labels: the use of text labels has been limited as much as possible.
• Interface testing: each major change of the interface has been discussed with the team of ISTC-CNR, developed, and double-checked with the team itself, until the appearance of SWift was considered entirely functional to its purpose; it is to underline that the team of ISTC-CNR includes deaf researchers, therefore a true sample of the main target users.

Buttons deserved particular attention. Each one is linked to an animation that demonstrates the effect of the action it performs, in order to support an easy grasp of its function. Although the application has been designed in strict contact with deaf users, we are planning more extensive testing to better analyze its usability and usefulness.

---

[2] available at:
<http://www.signbank.org/SignPuddle1.5/>

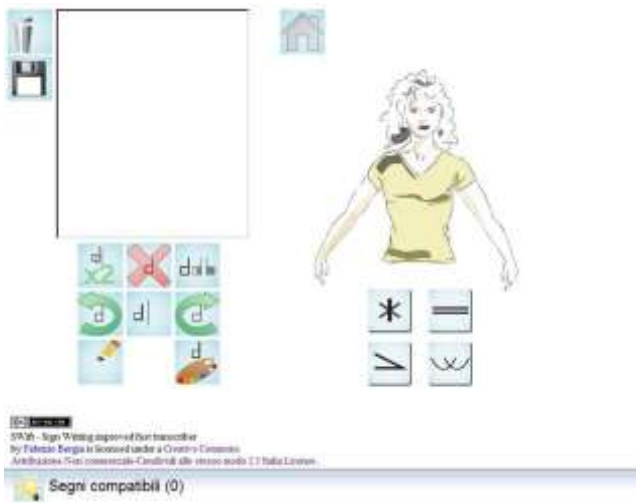

Fig. 5 – Home screen of SWift.

Choosing one body part, e.g. the hand, a number of boxes permit to choose the features of the specific sign, in this case the number of fingers, the visible side, etc. One choice is allowed for each box, and while choices are performed the set of available glyphs is modified accordingly (Fig. 6).

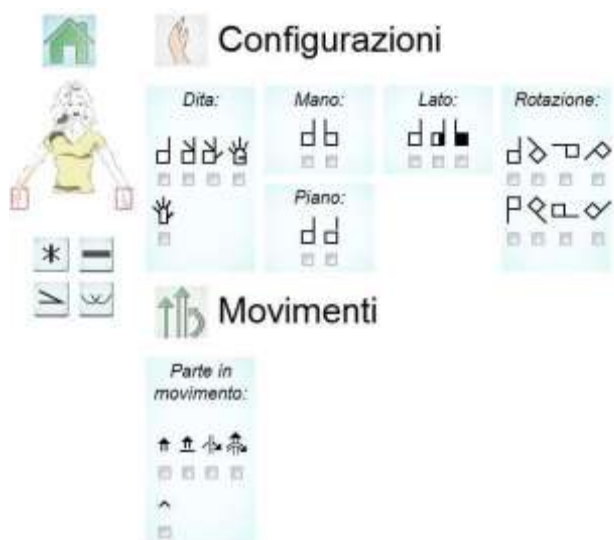

Fig. 6 – Glyph Menu - Search menu for the hands.

Glyphs are chosen in sequence until the desired sign is obtained. The sign can be stored for future reference; signs are stored together with the list of their components, therefore this form of memorization allows computing statistics and studying specific patterns and rules for the composition of glyphs, a great help for linguistic research.

## 5. Usability evaluation for SWift

### 5.1 Spatial Setting

We decided to asses SWift's capabilities and usability choosing the well-known "Think-Aloud Protocol", which – given our particular user community – had to be adapted in order to "fit" to deaf users: we can therefore define it a "Sign-Aloud Protocol". Before building up the test we found out that some steps had already been taken in that way: in fact, Roberts and Fels (2005) suggested a spatial setting to perform Think-Aloud-based usability tests with deaf people, where the environment is recorded by two cameras:

• CAM1 records the participant (rear view), the computer screen and the interpreter;
• CAM2 records the participant (front view) and the investigator.

The problem of using two cameras is the need to analyze two recordings at once instead of one, and to sync the data obtained by the two different sources Furthermore, when dealing with two separate videos, it would be difficult to maintain a synoptic view of everything that happens in the environment at any given time. After all, CAM 1 is only used in one of the experiments described in Roberts and Fels (2005), to obtain additional data from the participant. Bianchini (2012) proposed a system to perform the test using a single camera, introducing the use of a projector (Fig. 7 and Fig. 8).

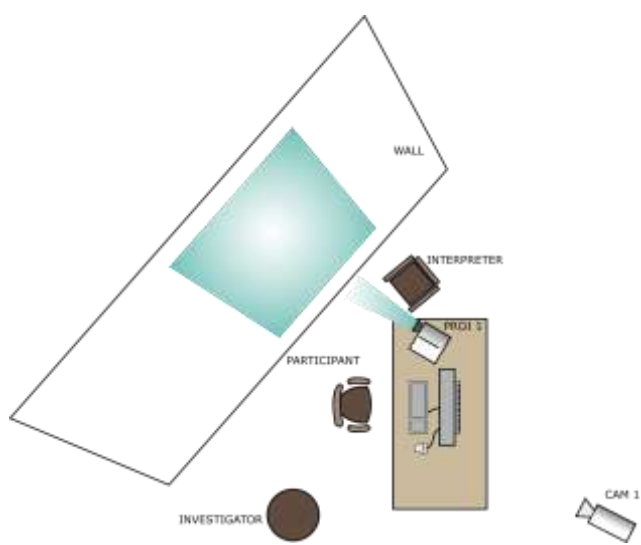

Fig. 7 - Spatial setting devised for SWift's evaluation (sketch)

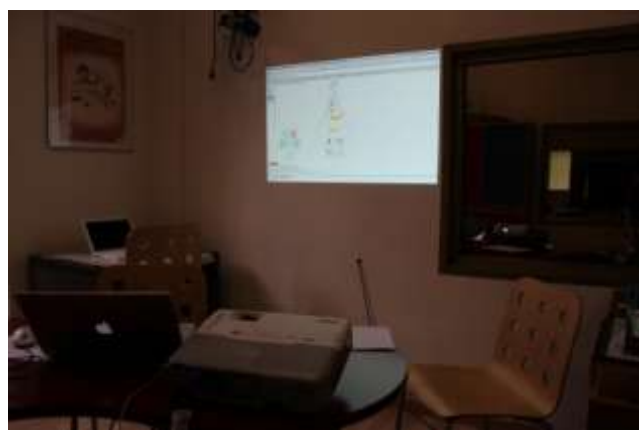

Fig. 8 - Spatial setting devised for SWift's evaluation (photo)

This way, CAM1 records everything:
- PARTICIPANT (front view)
- INTERPRETER
- COMPUTER SCREEN (projected on the wall)
- (INVESTIGATOR) (might not be necessary)

The camera oblique position (and thus the projector) with respect to the wall of the room is intended to minimize the obtrusion due to the monitor. The brightness of the room needs to be properly calibrated to allow optimal viewing of both the screen being projected on the wall and the signs produced by the participant and by the interpreter.

## 5.2 Test structure

The test is composed by three phases:

<u>The welcome time</u> - the participant faces a screen containing a signed video (on the left) and its transcription on the right. The informational payload of the video contains a thanksgiving for the participation, a brief explanation about the test structure and rules. The video also contains a reminder about the purpose of the test, which is not conceived to test the participant's ability at all, but to test the program capabilities and usability; this should help the participant to feel at ease with the test.

<u>The Sign-Aloud test</u> - this moment is the most important part of the procedure. The participant is required to perform a list of tasks to test the capabilities and the usability of SWift. In compliance with the Think-Aloud Protocol, the participant is asked to sign anything that comes to his/her mind. The user will need a reminder to recall which task is being carried out, what needs to be done, etc. This task list should be accessible both in VL and in SL. Among the available options devised, the most valid one is involving an interpreter that will always provide tasks translation in SL at the beginning of each task. This has the advantage of being non-invasive and ensuring a complete understanding of the tasks. The VL task list is proposed as a list of brief, simple questions addressed directly to the participant.

The required tasks have different difficulty levels. The test is designed to prevent user stress by alternating complex and simple tasks, ranging from the composition of a whole sign (complex) to the insertion of a user-chosen glyph (simple).

<u>The final usability questionnaire</u> – the participant are asked to answer to a usability questionnaire both in SL and in VL (Italian in this case) designed adapting the QUIS usability questionnaire (Slaughter *et al*. 1995) to the needs of deaf users. The questionnaire will stimulate the participants to express their thoughts and feelings about SWift.

## 5.3 Test results

The Think-Aloud protocol applied to deaf users has proven very useful to evaluate SWift's usability. During the first minutes of the test, the users were inclined to focus on resolving the tasks rather than signing their thoughts. It was necessary to remind them to express their thoughts very often. After some minutes, they complied with the prescribed rules.

No bugs leading to unpredictable program behavior and/or errors were found during the test, confirming SWift good design and bug testing. Anyway, some aspects need to be deeply investigated to get SWift perfectly operational. In particular, the statistical processing of test data highlighted some interesting trends. We noticed that the most recurring errors were strongly connected to the correct understanding of SWift core resource: the glyph search mechanism. More specifically, many users were inclined to ignore it (and its function) after seeing the first resulting glyphs. It is pretty straightforward, then, that the task with more errors is one of the earliest, the one in which the user interacts with the glyph search engine for the first time. The intuitiveness of the interface and the thorough design process have been rewarded by the excellent result of the tasks requiring the user to locate function activation points (involving icon decoding and recognition) but the use of graphics in the glyph search engine has proven difficult to understand for some users. As most issues were related to the glyph menu, some were affecting the reachability of some glyph families in the menu, causing sporadic episodes of disorientation during the search of very particular glyphs.

## 6. Conclusions

The SWift has all the necessary features to become a widely-used SW editor. The application was designed in close contact with the research team ISTC-CNR, and has been built to meet the needs of the deaf. While its validity will soon be tested in the field more extensively, the initial results are encouraging.

Much remains to be done: SWift is an editor, but the database has already been programmed so that, with appropriate modifications to the application business logic, it can offer more advanced features: it could lay the groundwork for a real SW text editor, or a converter between different ISWAs.

Two development lines have been identified for the SWift: one will lead to the replacement of the current glyph search engine with a OCR-like recognition engine, allowing the user to draw "free-hand" a glyph, which will be analyzed and recognized by the system and finally replaced with its "official" ISWA version. The second line will upgrade SWift with a semi-automatic image-processing engine allowing the users to produce signs in front of a web-cam and get a real-time transcription. This line of research will proceed step-by-step: first, a little help will be required from the users (e.g., showing the configuration before producing the sign, and so on) and then we will address the development of a fully automated transcription engine.

## 7. Acknowledgements

The present research work was partly funded by "VISEL: Vision, Deafness, Signs and E-learning - a bridge of letters and signs towards a knowledge society", a project realized within the framework of the funding for basic research (FIRB) allocated by the Italian Ministry of Education, University and Research (MIUR) [see: < http://www.visel.cnr.it >].

The ISTC-CNR (Institute of Cognitive Sciences and Technologies) provided research material, knowledge and qualified personnel to sustain and improve the present research work [see: < http://www.istc.cnr.it >].

In memory of Elena Antinoro Pizzuto.